\documentclass[10pt,twocolumn,letterpaper]{article}

\usepackage{iccv}
\usepackage{times}
\usepackage{epsfig}
\usepackage{graphicx}
\usepackage{amsmath}
\usepackage{amssymb}

\usepackage{booktabs}
\usepackage{subcaption}
\usepackage{float}
\usepackage{multirow}
\usepackage{multicol}
\usepackage{makecell}
\usepackage{pifont}
\usepackage{bbm}
\def\cross{\ding{55}}

\usepackage[accsupp]{axessibility}

\usepackage[breaklinks=true,bookmarks=false]{hyperref}

\iccvfinalcopy % *** Uncomment this line for the final submission

\def\iccvPaperID{****} % *** Enter the ICCV Paper ID here

\begin{document}

%%%%%%%%% TITLE

\title{Point-Query Quadtree for Crowd Counting, Localization, and More}

\author{Chengxin Liu$^1$ \quad Hao Lu$^1$ \quad Zhiguo Cao$^1$\thanks{corresponding author} \quad Tongliang Liu$^2$\\
		$^1$
        Key Laboratory of Image Processing and Intelligent Control, Ministry of Education\\
        School of Artificial Intelligence and Automation\\ 
        Huazhong University of Science and Technology, China\\
		$^2$The University of Sydney, Australia\\
		{\tt\small \{cx\_liu, hlu, zgcao\}@hust.edu.cn}
	}

\maketitle

%%%%%%%%% ABSTRACT

\begin{abstract}

We show that crowd counting can be viewed as a decomposable point querying process. This formulation enables arbitrary points as input and jointly reasons whether the points are crowd and where they locate. 
The querying processing, however, raises an underlying problem on the number of necessary querying points. Too few imply underestimation; too many increase computational overhead. To address this dilemma, we introduce a decomposable structure, i.e., the point-query quadtree, and propose a new counting model, termed Point quEry Transformer (PET). PET implements decomposable point querying via data-dependent quadtree splitting, where each querying point could split into four new points when necessary, thus enabling dynamic processing of sparse and dense regions. Such a querying process yields an intuitive, universal modeling of crowd as both the input and output are interpretable and steerable. We demonstrate the applications of PET on a number of crowd-related tasks, including fully-supervised crowd counting and localization, partial annotation learning, and point annotation refinement, and also report state-of-the-art performance. For the first time, we show that a single counting model can address multiple crowd-related tasks across different learning paradigms. Code is available at \url{https://github.com/cxliu0/PET}.

\end{abstract}

%%%%%%%%% INTRODUCTION

% Figure: concept
\begin{figure*}[ht]
	\centering
	\includegraphics[width=1.0\linewidth]{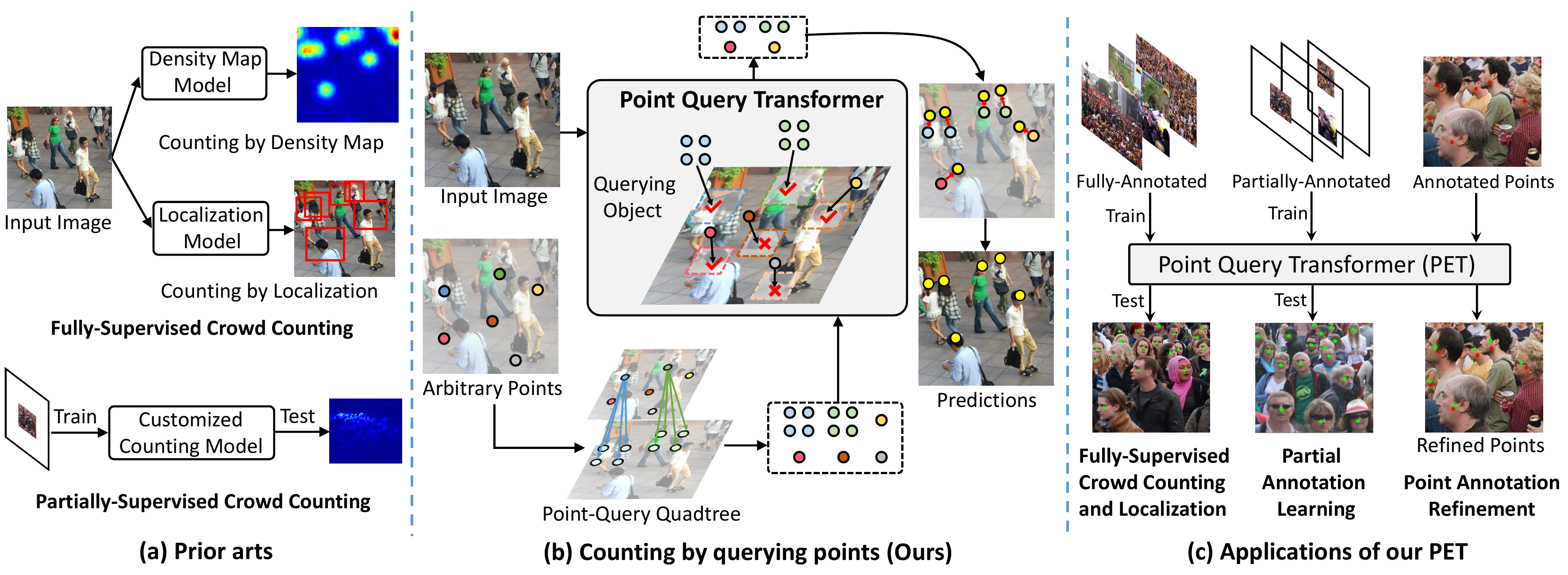}
	\caption{\textbf{Comparison between prior arts and our point-query counting paradigm}. In contrast to (a) prior arts, we consider arbitrary points as input, and reason whether each point is a person and where the person locates.	We devise (b) a point-query quadtree to deal with dense crowd with adaptive tree splitting. The query design renders PET an intuitive and universal approach, enabling (c) various applications, such as fully-supervised crowd counting and localization, partial annotation learning, and point annotation refinement.
	}
	\label{fig:fig1_concept}
\end{figure*}

\section{Introduction}

Crowd counting aims to estimate the number of crowd from an image. Existing approaches typically address counting by learning surrogate targets such as density maps, where the count is acquired by integrating the inferred density map. Despite being effective, they cannot provide an intuitive understanding of the crowd, \ie, no instance-level information is provided, which impedes high-level crowd analysis. Instead of merely predicting a count value from an image, some approaches focus on estimating the fine-grained information of crowd by either a head bounding box~\cite{liu2019box,sam2020box} or a single head point~\cite{cheng2021d2c,liu2019recurrent,song2021p2p}.
The former casts crowd counting as a head detection problem.
However, the detection accuracy has no guarantee in theory due to the lack of box information. In contrast, the latter directly outputs the head points, bypassing the error-prone stage of bounding box estimation. Nevertheless, they often require post-processing~\cite{cheng2021d2c,liu2019recurrent} to obtain the location of each person. As a result, congested scenes may render failures of counting or localization. In addition, prior arts typically tackle a specific counting task or a learning paradigm; each requires a customized design. This impedes their use in different applications or tasks. For example, a fully-supervised counting model often cannot well address semi-supervised crowd counting~\cite{xu2021partial}.

In this work, we formulate crowd counting as a decomposable point querying process. The point querying design allows a model to receive arbitrary points as input and to reason whether each point is a person and where it locates. 
An appealing property of this design is that it provides an intuitive and universal modeling of crowd. To be specific, the intuition lies in that each querying point physically corresponds to a person or background.
The arbitrariness of querying points implies that the position and the number of input points are both steerable. Therefore, by simply adjusting the input, our formulation naturally fits different crowd-related tasks, such as fully-supervised crowd counting and localization, partial annotation learning~\cite{xu2021partial}, and point annotation refinement (Fig.~\ref{fig:fig1_concept}c).

However, since an input image may contain an arbitrary number of crowd, it is non-trivial to predefine the number of querying points. 
In practice, too few points lead to underestimation, while too many points yield a large computational cost. To tackle this pitfall, 
we present a decomposable structure---point-query quadtree.
The key advantage of the quadtree is that it allows data-dependent splitting,
where one querying point could split into several new ones when necessary,
hence enabling dynamic processing of sparse and dense regions. Based on the quadtree, we instantiate a Point quEry Transformer (PET) to achieve decomposable point querying, as shown in Fig.~\ref{fig:fig1_concept}b. Another key ingredient of PET is the progressive rectangle window attention, where the querying process is performed within a local window rather than the whole image in a progressive manner for efficient inference. 

Extensive experiments on four crowd-counting benchmarks show that PET exhibits many appealing properties: i) \textit{generic}: PET is applicable to several crowd-related tasks, such as fully-supervised crowd counting and localization, partial annotation learning, and point annotation refinement;
ii) \textit{effective}: PET reports state-of-the-art crowd counting and localization performance against recent approaches. In particular, it achieves a mean absolute error (MAE) of $49.34$ on the ShanghaiTech PartA~\cite{zhang2016shab} dataset; 
iii) \textit{intuitive}: PET can proceed with the point query that physically corresponds to an object or background, and outputs also interpretable points.

The contributions of this work include the following:
\begin{itemize}
    \setlength\itemsep{0.1em}
    \item We show that decomposable point querying can be a universal crowd modeling idea to potentially unify crowd-related tasks.     
    
    \item We present PET, a Point Query Transformer for crowd counting, featured by the point-query quadtree and progressive rectangle window attention.
\end{itemize}

% Figure: pipeline 
\begin{figure*}[ht]
	\centering
	\includegraphics[width=\linewidth]{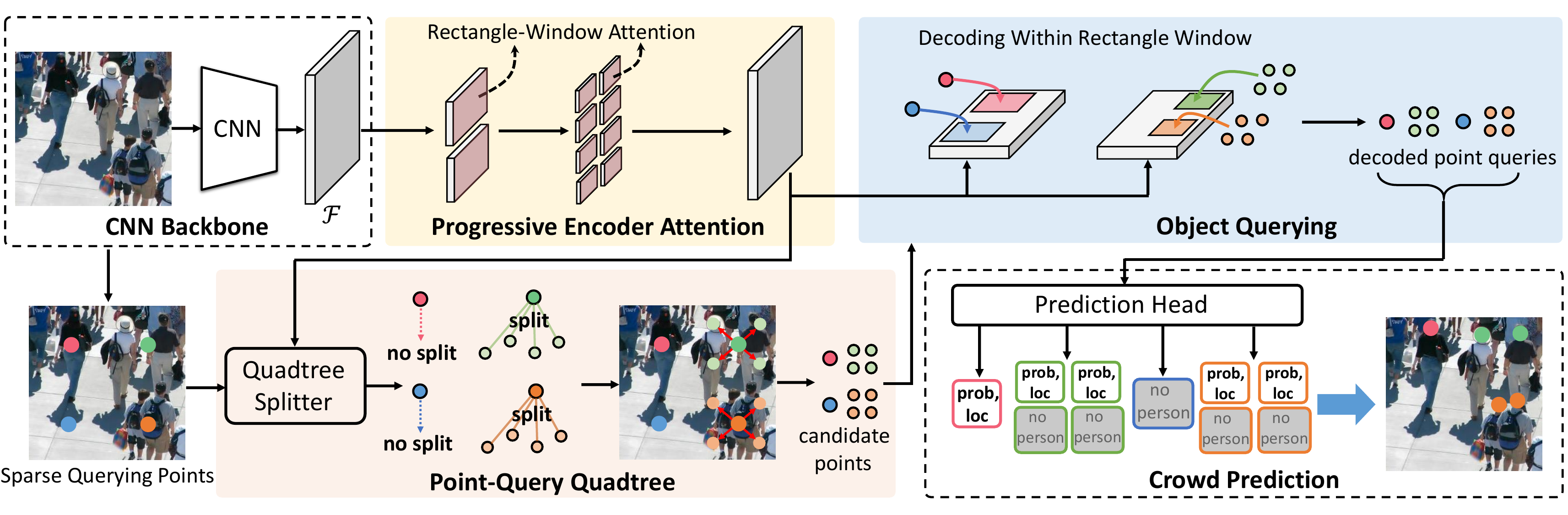}
	\caption{\textbf{Overall architecture of PET}. We first use a CNN backbone to extract the image representation $\mathcal{F}$. A transformer encoder with progressive rectangle window attention is then applied to $\mathcal{F}$ to encode context. Subsequently, the quadtree splitter receives sparse querying points and encoded features as input, outputting a point-query quadtree. The transformer decoder then decodes the point queries in parallel, where attention is computed within a local window. These point queries are finally passed through a prediction head to obtain crowd predictions, \ie, `no person' or `a person' with its probability and localization.} 
	\label{fig:pipeline}
	\vspace{-5pt}
\end{figure*}

%%%%%%%%% RELATED WORK

\section{Related Work}

We divide existing work according to whether they generate instance-level information of crowd, \ie, head bounding boxes or head points. In addition, we also discuss the recent progress of transformer-based approaches.

\vspace{-10pt}
\paragraph{Counting by Density Map.}
The majority of state-of-the-art approaches use density maps~\cite{victor2010density} as surrogate learning targets, where the count value is computed by integrating the predicted density map. These approaches advance the progress of crowd counting from various aspects, such as improving loss functions~\cite{ma2019bl,wan2020noisy,wang2020dm}, dealing with perspective effects~\cite{bai2020pers,yan2019pers,yang2020pers}, and exploiting contextual information~\cite{liu2019can}. To alleviate the inconsistency between the density map and the counting value, another line of work adopts the local counting paradigm~\cite{liu2020tcsvt,liu2020eccv,xiong2019sdcnet,xiong2023ijcv}, by classifying the count value of patches into discrete bins. Albeit successful, density map based approaches typically do not provide instance-level prediction.
In contrast, our approach can output the location of each person with a point, 
enabling a more intuitive understanding of crowd.

\vspace{-10pt}
\paragraph{Counting by Localization.}
Instead of predicting an intermediate representation like density maps, another alternative is to simultaneously estimate the count and location of crowd. Such fine-grained estimation of crowd has received substantial interest in recent years. For example, some recent approaches~\cite{lian2019loc,liu2019box,sam2020box} cast counting as a head detection problem, predicting the bounding boxes of heads. However, the pseudo ground-truth boxes generated from weak point supervision are error-prone, especially in congested regions. As a result, it not only hinders model training but also leads to inaccurate box predictions. Beyond bounding boxes, several approaches~\cite{cheng2021d2c,liu2019recurrent,song2021p2p} directly estimate the head points. However, these approaches often need heuristic post-processing to identify individual crowd.
Different from previous methods, our approach receives arbitrary points as input and predicts crowd by explicitly modeling the relation between a point and its surroundings. It streamlines the prediction process and enables applications in other crowd-related tasks.

\vspace{-10pt}
\paragraph{Transformer Based Counting.}
The recent success of vision transformers~\cite{nicolas2020detr,alexey2021vit} has sparked their applications in various computer vision tasks. 
Recently, much effort~\cite{gao2021swincrowd,liang2022cltr,lin2022man,sun2021bcct,ye2021cctrans,wei2021scene} 
has been devoted to deploying transformer architectures to crowd counting. Due to the strong representation capability of transformer, existing approaches mainly focus on developing a strong transformer backbone for feature extraction and incorporating it with a prediction module to estimate the count value. 
In contrast, we present a new formulation for crowd counting and instantiate it with a customized point query transformer.

%%%%%%%%% METHOD

\section{Counting Crowd by Querying Points}

\subsection{Problem Formulation}

In this work, we formulate crowd counting as a decomposable point querying process.
During the querying process,
sparse points could split into new points when necessary such that dense regions can be processed adaptively. In this way, counting can be achieved by querying those (split) input points.

We embody this formulation with a customized Point quEry TRansformer (PET). In PET, two ingredients are essential: 
i) the design of the point-query quadtree; 
ii) a progressive rectangle window attention mechanism. 
The former adaptively generates querying points to tackle dense crowd predictions, and the latter improves efficiency. 

\subsection{Architecture Overview}
The overall architecture of PET is depicted in Fig.~\ref{fig:pipeline}. It includes four components: a CNN backbone, an efficient encoder-decoder transformer, a point-query quadtree, and a prediction head.

Given an input image, the CNN backbone first extracts image representation, outputting the feature $\mathcal{F} \in \mathbb{R}^{h\times w \times c}$. Each spatial element in $\mathcal{F}$ is treated as a \textit{token}, which results in $h\times w$ tokens.
These tokens are then passed through a transformer encoder, implemented with progressive rectangle window attention, to encode contextual information. 
The rationale behind this design is efficiency, because a large number of tokens render global attention computationally expensive, especially for high-resolution images.

To obtain an instance-level understanding of the crowd, we construct a point-query quadtree to query the crowd. 
The point-query quadtree allows data-dependent splitting in congested regions such that dense and sparse scenes can be processed adaptively.
By receiving the point-query quadtree and encoded features as input, 
the transformer decoder reasons the relation between point queries under the guidance of image context, and subsequently, decodes the point queries in parallel. These decoded point queries finally pass through a prediction head to acquire crowd predictions.

\subsection{Point-Query Quadtree}

The design of the point query is vital to the success of our formulation. Considering that an input image may contain an arbitrary number of crowd, it is irrational to use a fixed number of queries as in object detection~\cite{nicolas2020detr}.
To enable scalable crowd estimation, point query should be meaningful and flexible. 
This yields three problems: 1) how to adapt the number of querying points to different scenes; 2) how to represent a point query; 3) how to predict crowd via point queries. We address them as follows.

\begin{figure}[t]
\centering
    \includegraphics[width=\linewidth]{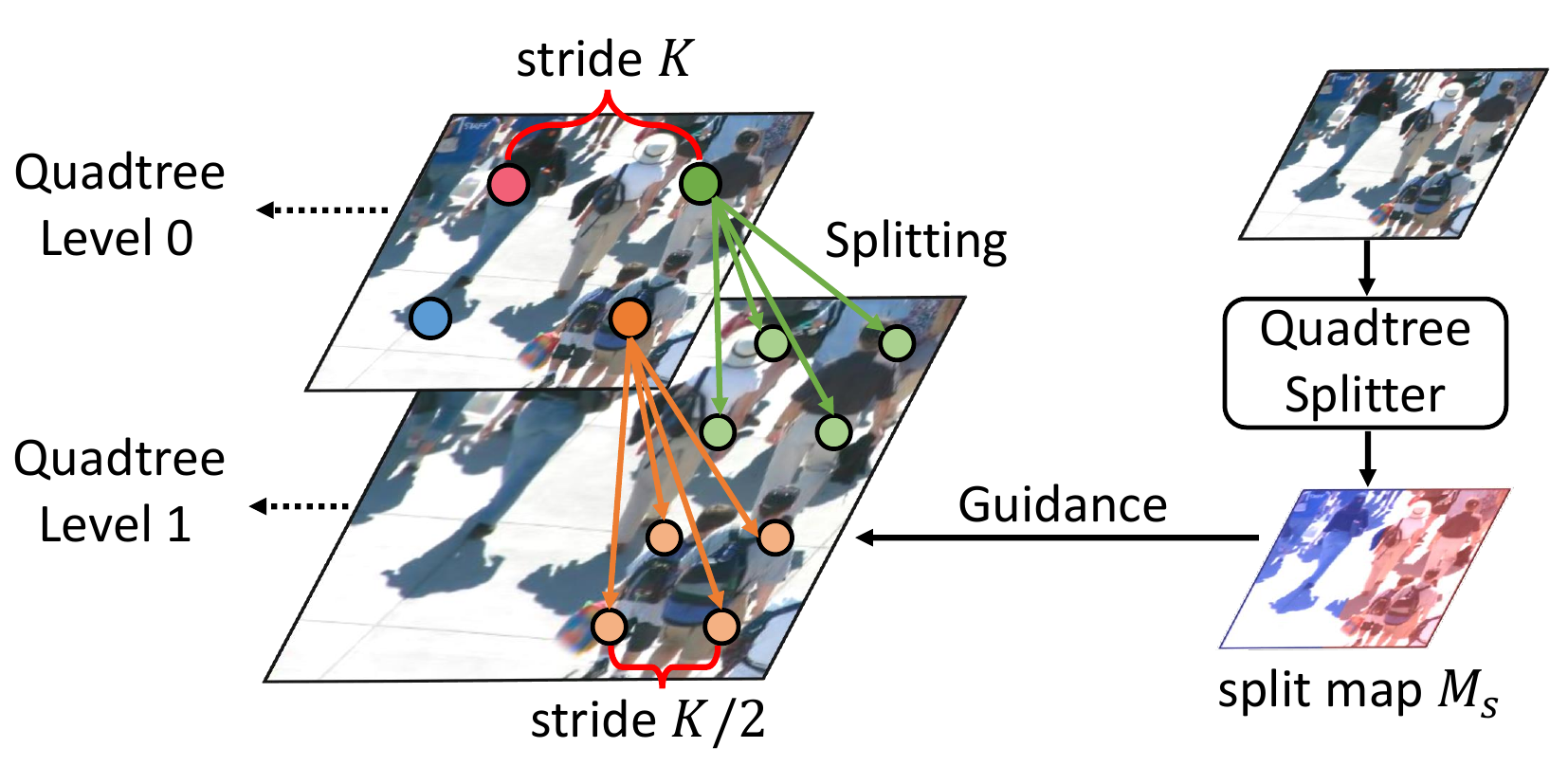}
    \caption{\textbf{Illustration of the point-query quadtree}. 
    The split map $M_s$ is upsampled to the original image size, 
    where red regions require quadtree splitting.
    }
    \label{fig:quadtree}
\end{figure}

\begin{figure}[t]
\centering
    \includegraphics[width=\linewidth,height=0.25\linewidth]{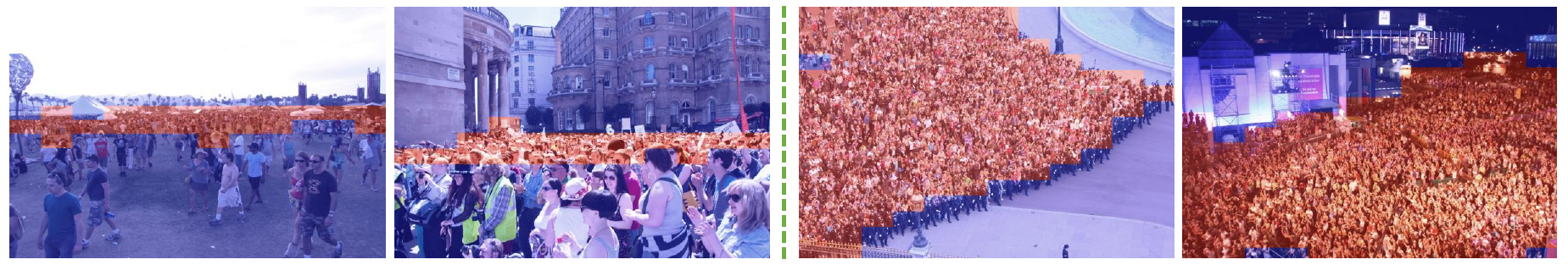}
    \caption{\textbf{Examples of the split map}. The red regions denote the congested regions that require  tree 
    splitting.}
    \label{fig:quadtree_splitter}
    \vspace{-10pt}
\end{figure}

\vspace{-5pt}
\paragraph{Quadtree Construction.}
To make point query scalable to sparse and dense scenes, we propose a decomposable structure, termed point-query quadtree.
The quadtree follows a sparse to dense process, \ie,
sparse querying points are first spanned across the image,
and they are adaptively split into dense querying points in congested scenes for dense crowd prediction. A criterion is thus necessary to decide when to split sparse querying points.
We consider that the splitting process should be determined by inspecting a local region, instead of relying on a single point. We therefore adopt a region-based quadtree splitter to construct the quadtree.

\textit{Quadtree Splitting.}
Fig.~\ref{fig:quadtree} illustrates the construction process of the point-query quadtree. 
Specifically, we first uniformly set sparse querying points on the image with a stride of $K$, which corresponds to the quadtree level $0$.
A quadtree splitter is then applied to the encoded features, outputting a split map $M_s \in \mathbb{R}^{h'\times w'}$. 
Each element in $M_s$ represents the probability of a region being dense, where $1$ denotes dense regions and $0$ denotes sparse ones. 
The initial querying points in dense regions are split into dense querying points, forming the quadtree level $1$. This process repeats until the maximum splitting time $L$ is reached, which results in a $L+1$-layer quadtree. 
A multi-layer quadtree could be implemented by using multiple quadtree splitters to predict the split map of each layer.
One may consider how many times the quadtree needs to split to tackle dense regions.
In practice, we find that splitting once ($L$$=$$1$) is generally sufficient to deal with crowd estimation. Please refer to the supplementary for a detailed analysis.

In particular, the quadtree splitter consists of an average pooling layer and a $1\times 1$ convolution layer followed by a $\tt sigmoid$ function, so the computational cost is negligible. We show some examples of the output of the quadtree splitter in Fig.~\ref{fig:quadtree_splitter}, in which red regions denote congested regions requiring quadtree splitting. We observe that the quadtree splitter can distinguish the congested regions. Note that the split map is binarized using a threshold of $0.5$.

\vspace{-5pt}
\paragraph{Point Query Representation.}
Given a querying point with pixel location $(x,y)$, we need to represent it as a point query. Our intuition tells us that a point query should contain both semantic and localization information. To encode the semantics of point query, we repurpose the CNN feature. Technically, we interpolate the CNN features $\mathcal{F}$ to the original image size and sample the feature vector $\mathcal{F}_{x,y} \in \mathbb{R}^{1\times 1 \times c}$. For position information, we adopt a fixed spatial positional embedding following~\cite{nicolas2020detr}. The positional embedding and $\mathcal{F}_{x,y}$ are summed to form the point query representation.

\vspace{-5pt}
\paragraph{Crowd Prediction.} The final step is how to predict crowd via point queries. Following the philosophy of dotted annotations (one dot per person), we represent each person by a unique point query.
Specifically, we first pass the point-query quadtree through the transformer decoder to obtain the decoded representation. A prediction head is then applied to the decoded point queries, outputting a set of predicted persons $\mathcal{Q}= \{\mathbf{q}_i\}_{i=1}^{N}$. Note that $\mathbf{q}_i$ consists of the classification probability $c_i \in [0,1]$ and the normalized pixel location $\mathbf{p}_i=(x_i+\Delta x_i, y_i+\Delta y_i)$, where $(x_i,y_i)$ is the pixel location of a point query and $(\Delta x_i,\Delta y_i)$ is the predicted offsets. The prediction head is composed of MLP layers with ReLU activation.

We remark that the prediction head receives a varied number of queries for different images,
as the queries are generated on-the-fly by the point-query quadtree. This ensures that sparse and dense point queries only operate on corresponding regions, avoiding unnecessary computation.

% Figure: horizontal vs. verticle
\begin{figure}[t]
	\centering
	\includegraphics[width=0.42\textwidth]{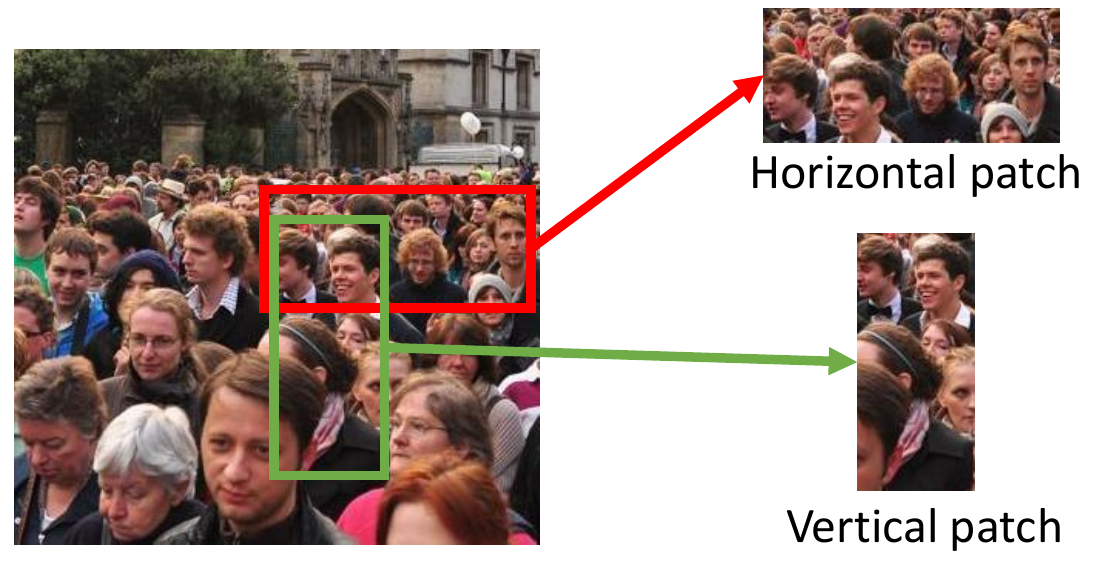}
	\caption{
	\textbf{The motivation of using horizontal window.} The horizontal patch often contains more people than the vertical one due to perspective prior.
	}
	\label{fig:horizontal}
\end{figure}

% Figure: rectangle window attention
\begin{figure}[t]
	\centering
	\includegraphics[width=0.48\textwidth]{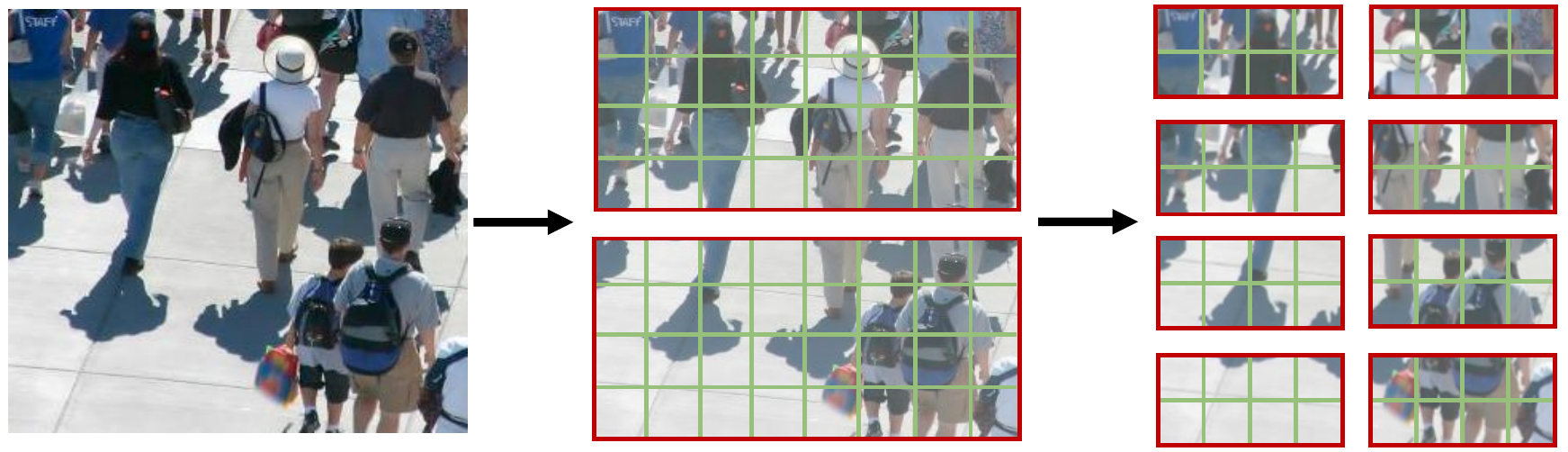}
	\caption{
	\textbf{Illustration of progressive encoder attention}. 
        }
	\label{fig:rect_attention}
	\vspace{-10pt}
\end{figure}

\subsection{Progressive Querying in Rectangle Window}
\label{sec:rect_win}

Here we delineate the design of our transformer encoder and decoder. In general, we perform object querying in a progressive manner within rectangle window.

\vspace{-5pt}
\paragraph{Progressive Encoder Attention.} 
To encode crowd information of different scales, we adopt a progressive attention mechanism. 
The idea is that the transformer encoder first inspects a sufficiently large region and then focuses on a small one. Considering that the perspective change often occurs in crowd images, we compute attention within a rectangle window. More specifically, we design a horizontal window since it usually contains more people than the vertical one (Fig.~\ref{fig:horizontal}). 

The idea of the progressive encoder attention is illustrated in Fig.~\ref{fig:rect_attention}.
Given an input image, we first compute attention within a relatively large rectangle window in the first few encoder layers. Each window is with a size of $s_e\times r_e s_e$, where $s_e$ is the height of the rectangle window, and $r_e$ is the aspect ratio. Then, a small rectangle window (with a size of $\frac{1}{2} s_e\times \frac{1}{2}r_e s_e$) is adopted to perform attention in the subsequent encoder layers.
The encoder attention is computed as:
\begin{align}
     \hat{\mathbf{x}}^{l} &= \texttt{LN}(\texttt{RectWin-SA}(\mathbf{x}^{l-1}) + \mathbf{x}^{l-1})\,,
     \nonumber \\
    \mathbf{x}^{l} &= \texttt{LN}(\texttt{FFN}(\hat{\mathbf{x}}^{l}) + \hat{\mathbf{x}}^{l})\,,
\end{align}
where $\mathbf{x}^{l-1}$ and $\mathbf{x}^l$ are the output features of the encoder layer $l-1$ and the layer $l$, respectively. Note that $\mathbf{x}^{0}$ is initialized with $\mathcal{F}$. 
\texttt{RectWin-SA}, \texttt{FFN}, and \texttt{LN} 
denote rectangle window self-attention, feed-forward network, and layer normalization, respectively. 
Benefiting from the design of progressive rectangle window attention, we can achieve efficient computation with linear complexity, which is particularly helpful when processing high-resolution images. 
Detailed architecture of the encoder can be found in the supplementary.

\vspace{-5pt}
\paragraph{Decoding Within Rectangle Window.}
The transformer decoder receives the point-query quadtree and encoded image features as input, outputting decoded point queries. It reasons crowd by modeling the relation between point queries under the guidance of image context. 
Intuitively, we decide whether a point query is a person based on its surrounding point queries and context. We therefore propose to compute decoder attention within local windows. 
Considering the hierarchy structure of quadtree, we also adopt progressive attention when decoding point queries.

Technically, the attention of sparse point queries is computed in a relatively large rectangle window (with a size of $\frac{1}{2} s_e\times \frac{1}{2} r_e s_e$), and the window size of dense point queries is reduced to $\frac{1}{4} s_e\times \frac{1}{4} r_e s_e$. 
In particular, the decoder attention is computed as:
\begin{align}
     \hat{\mathbf{z}}^{l} &= \texttt{LN}(\texttt{RectWin-SA}(\mathbf{z}^{l-1}) + \mathbf{z}^{l-1})\,, \nonumber \\
     \hat{\mathbf{z}}^{l} &= \texttt{LN}(\texttt{RectWin-CA}(\hat{\mathbf{z}}^{l}, \mathbf{x}^{N}) + \hat{\mathbf{z}}^{l})\,, \nonumber \\
    \mathbf{z}^{l} &= \texttt{LN}(\texttt{FFN}(\hat{\mathbf{z}}^{l}) + \hat{\mathbf{z}}^{l})\,,
\end{align}
where $\mathbf{x}^{N}$ is the final output of the transformer encoder, $\mathbf{z}^{l-1}$ and $\mathbf{z}^l$ are the output features of the decoder layer $l-1$ and the layer $l$, respectively. 
Note that $\mathbf{z}^0$ is initialized with the representation of point queries. 
\texttt{RectWin-SA} and \texttt{RectWin-CA} denote the rectangle window self-attention and the rectangle window cross-attention, respectively. 

% Table: Comparisons with SOTA
\begin{table*}[t]
    \centering
	\caption{Quantitative comparison of crowd counting results on the ShanghaiTech~\cite{zhang2016shab}, UCF-QNRF~\cite{haroon2018ucf}, and JHU-Crowd++~\cite{sindagi2020jhu} datasets. The best performance is in \textbf{boldface}, and the second best is \underline{underlined}.}
		\setlength{\tabcolsep}{6pt}
		\begin{tabular}{@{}l l c |l l| l l| l l| l l @{}}
			\toprule
			\multirow{2}{*}{Method} & \multirow{2}{*}{Venue} & \multirow{2}{*}{Localization} & \multicolumn{2}{c|}{SH PartA} & \multicolumn{2}{c|}{SH PartB} & \multicolumn{2}{c|}{UCF-QNRF} & \multicolumn{2}{c}{JHU-Crowd++}  \\
			& & & MAE & MSE & MAE & MSE & MAE & MSE & MAE & MSE \\
			 
			 \midrule
			 CSRNet~\cite{li2018csrnet} & CVPR'18 & \cross & 68.2 & 115.0 & 10.6 & 16.0 & - & - & 85.9 & 309.2 \\
			 CAN~\cite{liu2019can} & CVPR'19 & \cross & 62.3 & 100.0 & 7.8 & 12.2 & 107.0 & 183.0 & 100.1 & 314.0\\
			 BL+~\cite{ma2019bl} & ICCV'19 & \cross & 62.8 & 101.8 & 7.7 & 12.7 & 88.7 & 154.8 & 75.0 & 299.9\\
			 ASNet~\cite{jiang2020asnet} & CVPR’20 & \cross & 57.78 & 90.13 & - & - & 91.59 & 159.71 & - & - \\
			 DM-Count~\cite{wang2020dm} & NeurIPS’20 & \cross & 59.7 & 95.7 & 7.4 & 11.8 & 85.6 & 148.3 & - & - \\
			 NoisyCC~\cite{wan2020noisy} & NeurIPS’20 & \cross & 61.9 & 99.6 & 7.4 & 11.3 & 85.8 & 150.6 & 67.7 & 258.5 \\
			 SDA+DM~\cite{ma2021sda} & ICCV'21 & \cross & 55.0 & 92.7 & - & - & \underline{80.7} & 146.3 & \underline{59.3} & 248.9  \\
			 GauNet+CSRNet~\cite{cheng2022gaunet} & CVPR'22 & \cross & 61.2 & 97.8 & 7.6 & 12.7 & 84.2 & 152.4 & 69.4 & 262.4 \\
			 
			 \midrule
			 GL~\cite{wan2021gl} & CVPR'21 & \checkmark & 61.3 & 95.4 & 7.3 & 11.7 & 84.3 & 147.5 & 59.9 & 259.5 \\
			 P2PNet~\cite{song2021p2p} & ICCV'21 & \checkmark & \underline{52.74} & \underline{85.06} & \underline{6.25} & \underline{9.9} & 85.32 & 154.5 & - & - \\
			 CLTR~\cite{liang2022cltr} & ECCV'22 & \checkmark & 56.9 & 95.2 & 6.5 & 10.6 & 85.8 & \textbf{141.3} & 59.5 & \underline{240.6} \\
			 PET - Ours & - & \checkmark & \textbf{49.34} & \textbf{78.77} & \textbf{6.19} & \textbf{9.69} &\textbf{79.53} & \underline{144.32} & \textbf{58.5} & \textbf{238.0} \\
    	    \bottomrule
	    \end{tabular}
	\label{table:sota}
\end{table*}

\subsection{Network Optimization}

\paragraph{Training.} 
The output of PET is a set of candidate persons $\mathcal{Q}= \{\mathbf{q}_i\}_{i=1}^{N}$.
We optimize the network using bipartite matching~\cite{nicolas2020detr} between the predictions and ground truth points $\mathcal{Y}={\{\mathbf{y}_i\}_{i=1}^{M}}$.
The loss is computed as:
\begin{equation}
    \ell_{pq} = \frac{1}{N} \sum_{i=1}^{N} \ell_{cls} (c_i, c_i^*) +
    \lambda_1 \frac{1}{M} \sum_{i=1}^{M} \ell_{loc} (\mathbf{p}_{\sigma(i)}, \mathbf{y}_i)\,,
\end{equation}
where $i$ is the index of a point query, $c_i$ is the predicted classification probability, $\lambda_1$ is a hyperparameter, and $\sigma(i)$ denotes the index of point query that is matched with the ground-truth point $\mathbf{y}_i$. $\sigma$ is obtained by bipartite matching~\cite{nicolas2020detr}.
The classification label $c_i^*$ equals to $1$ only if the point query is matched with a ground-truth point, and $0$ otherwise.
For the classification loss $\ell_{cls}$ and localization loss $\ell_{loc}$, we employ cross entropy loss and smooth $\ell_1$ loss~\cite{ross2015sl1}.
In addition, the quadtree splitter is supervised by:
\begin{equation}
\label{eq:split}
    \ell_{split} = \mathbbm{1}(\texttt{dense}) (1-\max(M_s)) + \min(M_s)\,,
\end{equation}
where ${M_s}$ is the split map, and $\mathbbm{1}(\texttt{dense})$ equals $1$ if the input image has dense regions, otherwise $0$. We consider an image has dense regions if its crowd density is high. 
Recall that each element in ${M_s}$ represents the probability of a region being dense.
Eq.~\eqref{eq:split} functions as a minimum supervision for the quadtree splitter to discriminate dense and sparse regions.
We retain the second term as sparse regions generally exist in an image.
The final loss function is:
\begin{equation}
    \ell_{total} = \ell_{pq} + \lambda_2 \ell_{split}\,,
\end{equation}
where $\lambda_2$ is a weight-balancing hyperparameter.

\textit{Dual Supervision.} 
To prevent the loss from being diluted by samples with few people, we introduce dual supervision.
Given a mini-batch of samples, we divide them into sparse and dense packs according to crowd density, and separately compute $\ell_{total}$ on these two packs. The losses are then summed for backpropagation.

\vspace{-5pt}
\paragraph{Inference.} During testing, the point-query quadtree is dynamically constructed based on the split map $M_s$. The final predictions are obtained by thresholding the classification probability of point queries, \eg, a threshold of $0.5$.

%%%%%%%%% EXPERIMENTS

\section{Results and Discussion}

\subsection{Datasets and Implementation Details} \label{sec:implementation}

\paragraph{Datasets.}
We evaluate our approach on four crowd counting datasets, including ShanghaiTech~\cite{zhang2016shab}, UCF-QNRF~\cite{haroon2018ucf}, JHU-Crowd++~\cite{sindagi2020jhu}, and NWPU-Crowd~\cite{wang2021nwpu}. Following previous work~\cite{li2018csrnet,zhang2016shab}, we use mean absolute error (MAE) and mean square error (MSE) as the evaluation metrics.

ShanghaiTech~\cite{zhang2016shab} has two subsets, including PartA ($300/182$ images for train/test) and PartB ($400/316$ images for train/test). UCF-QNRF~\cite{haroon2018ucf} contains $1535$ images with diverse crowd density ($1201/334$ for train/test). JHU-Crowd++~\cite{sindagi2020jhu} is a large-scale crowd counting dataset with $4372$ images ($2272$/$500$/$1600$ images for train/val/test), which covers various scenarios. NWPU-Crowd~\cite{wang2021nwpu} is also a large-scale dataset, which contains $3109$, $500$, and $1500$ images for training, validation, and testing, respectively.

\paragraph{Implementation Details.} PET is optimized using the Adam optimizer~\cite{ilya2019adam} with the weight decay of $5\times 10^{-4}$. The initial learning rate of $10^{-5}$ is set for the CNN backbone 
(we use VGG16~\cite{karen2015vgg}), and $10^{-4}$ for the transformer. 
The point-query quadtree has a maximum depth of $2$, and the initial stride of sparse point queries is set to $K=8$. The layer number of transformer encoder and decoder is $4$ and $2$, respectively. Note that our point-query quadtree shares the same decoder. We set window parameters as $s_e=16$ and $r_e=2$. For loss coefficients, we set $\lambda_1=5.0$ and $\lambda_2=0.1$ to balance the contribution of different terms. 

Regarding data augmentation, we randomly crop $256\times256$ patches from each image as training samples, and perform random scaling and random flipping.
For datasets that contain high-resolution images, we resize the image and keep the original aspect ratio. The longer side of each image is constrained within $1536$, $2048$, and $2048$ pixels for UCF-QNRF, JHU-Crowd++, and NWPU-Crowd, respectively.

% Table: NPWU-Crowd 
\begin{table*}[!htb]
    \centering
    \caption{Crowd counting and localization results on the NWPU-Crowd~\cite{wang2021nwpu} test set. }
    \begin{subtable}{.35\linewidth}
      \centering
        \caption{Crowd counting results.}
        \begin{tabular}{@{}l l l l @{}}
    		\toprule
    		Method & Venue &  MAE & MSE  \\
    			 
    		\midrule
    		CSRNet~\cite{li2018csrnet} & CVPR'18 & 121.3 & 387.8 \\
    		BL+~\cite{ma2019bl} & ICCV'19 &  105.4 & 454.2 \\
    		S-DCNet~\cite{xiong2019sdcnet} & ICCV'19 & 90.2 & 370.5 \\
    		NoisyCC~\cite{wan2020noisy} & NeurIPS'20 & 96.9 & 534.2 \\
    		DM-Count~\cite{wang2020dm} & NeurIPS'20 & 88.4 & 388.6 \\
    		P2PNet~\cite{song2021p2p} & ICCV'21 & 77.44 & 362 \\
    	    MAN~\cite{lin2022man} & CVPR'22 & \underline{76.5} & \textbf{323.0} \\
    		\midrule
    		PET - Ours & - & \textbf{74.4} & \underline{328.5} \\
    	    \bottomrule
    	\end{tabular}
    	\label{table:nwpu}
    \end{subtable}%
    \hspace{35pt}
    \begin{subtable}{.55\linewidth}
      \centering
        \caption{Crowd localization results. \textbf{F1-m/Prec/Rec}: F-measure/precision/recall.}
        \begin{tabular}{@{}l l l l l l l@{}}
		  \toprule
			\multirow{2}{*}{Method}  & \multicolumn{3}{c}{$\sigma_{l}$ (large threshold)} & \multicolumn{3}{c}{$\sigma_{s}$ (small threshold)} \\
			& F1-m & Prec & Rec & F1-m & Prec & Rec  \\
			\midrule
			Faster RCNN~\cite{ren2017faster} & 0.067 & \textbf{0.958} & 0.035 & 0.063 & \textbf{0.894} & 0.033 \\
			TinyFaces~\cite{hu2017tiny} & 0.567 & 0.529 & 0.611 & 0.526 & 0.491 & 0.566 \\
			RAZNet~\cite{liu2019recurrent} & 0.599 & 0.666 & 0.543 & 0.517 & 0.576 & 0.470 \\
			GL~\cite{wan2021gl} & 0.660 & \underline{0.800} & 0.562 & 0.587 & \underline{0.711} & 0.500 \\
			D2CNet~\cite{cheng2021d2c} & \underline{0.700} & 0.741 & 0.662 & \underline{0.632} & 0.670 & \underline{0.598} \\
			CLTR~\cite{liang2022cltr} & 0.685 & 0.694 & \underline{0.676} & 0.591 & 0.599 & 0.583 \\
    		\midrule
    		PET - Ours & \textbf{0.742} & 0.752 & \textbf{0.732} & \textbf{0.675} & 0.684 & \textbf{0.666} \\
    	    \bottomrule
	    \end{tabular}
	    \label{table:nwpu_loc}
    \end{subtable} 
\end{table*}

\subsection{Main Results} 
\label{sec:experiment}

The point query design of PET makes it applicable to several crowd-related tasks. In this section, we demonstrate three applications, including fully-supervised crowd counting and localization, partial annotation learning, and point annotation refinement. 

\vspace{-10pt}
\paragraph{Comparison With State of the Art.}
We compare PET with state-of-the-art methods on four datasets, in the context of fully-supervised crowd counting and localization.

Table~\ref{table:sota} reports the crowding counting results on ShanghaiTech~\cite{zhang2016shab}, UCF-QNRF~\cite{haroon2018ucf}, and JHU-Crowd++~\cite{sindagi2020jhu}. Our method not only achieves state-of-the-art results but also outputs localization information. In particular, PET significantly outperforms existing methods on ShanghaiTech PartA, with an MAE of $49.34$. The good performance on UCF-QNRF and JHU-Crowd++ also supports the adaptation of PET to dense scenes, because these datasets contain images with extremely dense crowd.

For the NWPU-Crowd~\cite{wang2021nwpu} dataset, the crowd counting results in Table~\ref{table:nwpu} show that PET performs favorably against state-of-the-art methods. It is worth noticing that PET outperforms existing localization-based approaches by a considerable margin, as indicated in Table~\ref{table:nwpu_loc}.
This validates the localization accuracy of PET.

% Table: partial annotation Learning
\begin{table}[t]
    \centering
	\caption{Quantitative results of partial annotation learning on the ShanghaiTech dataset. Ratio denotes the proportion of annotated region in each image. F. S. stands for fully-supervised and P. A. for partial annotation.}
		\setlength{\tabcolsep}{4pt}
		\begin{tabular}{@{}l l l l l l l@{}}
			\toprule
			\multirow{2}{*}{Method} & \multirow{2}{*}{Type} &  \multirow{2}{*}{Ratio} & \multicolumn{2}{c}{SH PartA} & \multicolumn{2}{c}{SH PartB} \\
			& & & MAE & MSE & MAE & MSE  \\
			\midrule
			MCNN~\cite{zhang2016shab} & F. S. & 100\% & 110.2 & 173.2 & 26.4 & 41.3    \\
			CSRNet~\cite{li2018csrnet} & F. S. & 100\% & 68.2 & 115.0 & 10.6 & 16.0 \\
			BL+~\cite{ma2019bl}  & F. S. & 100\% & 62.8 & 101.8 & 7.7 & 12.7 \\
			\midrule
			 Xu \textit{et al.}~\cite{xu2021partial} & P. A. & 10\% & \underline{72.79} & \underline{111.61} & \underline{12.03} & \textbf{18.70} \\
			 PET - Ours & P. A. & 10\% & \textbf{60.36} & \textbf{104.13} & \textbf{10.75} & \underline{19.40}\\
    	    \bottomrule
	    \end{tabular}
	\label{table:partial}
\end{table}

\vspace{-5pt}
\paragraph{Results on Partial Annotation Learning.} 
Learning with partial annotations~\cite{xu2021partial} is a new problem setting in crowd counting, in which only a partial region of each image is annotated. It aims to reduce the annotation cost and leverages data captured under various scenes. PET is applicable to this task because it can infer unlabelled regions based on the annotated ones, owing to the design of the point query. In particular, we follow the setting of~\cite{xu2021partial} and simply adopt a two-step training process. We first train PET with partial annotations, then we infer the annotations around annotated regions and retrain PET with a fusion of ground-truth points and inferred annotations. In contrast to~\cite{xu2021partial}, we do not require a specifically designed consistency criterion to constrain model training. Table~\ref{table:partial} shows the quantitative results on the ShanghaiTech dataset. Despite the simplicity of our training process, our approach outperforms~\cite{xu2021partial} by a large margin, especially on SH PartA. 
In addition, PET also performs favorably against fully-supervised methods.
This validates the effectiveness of PET on limited data and also the universal property of point-query modeling.

% Figure: Annotation Refinement
\begin{figure}[t]
	\centering
        \includegraphics[width=0.65\linewidth]{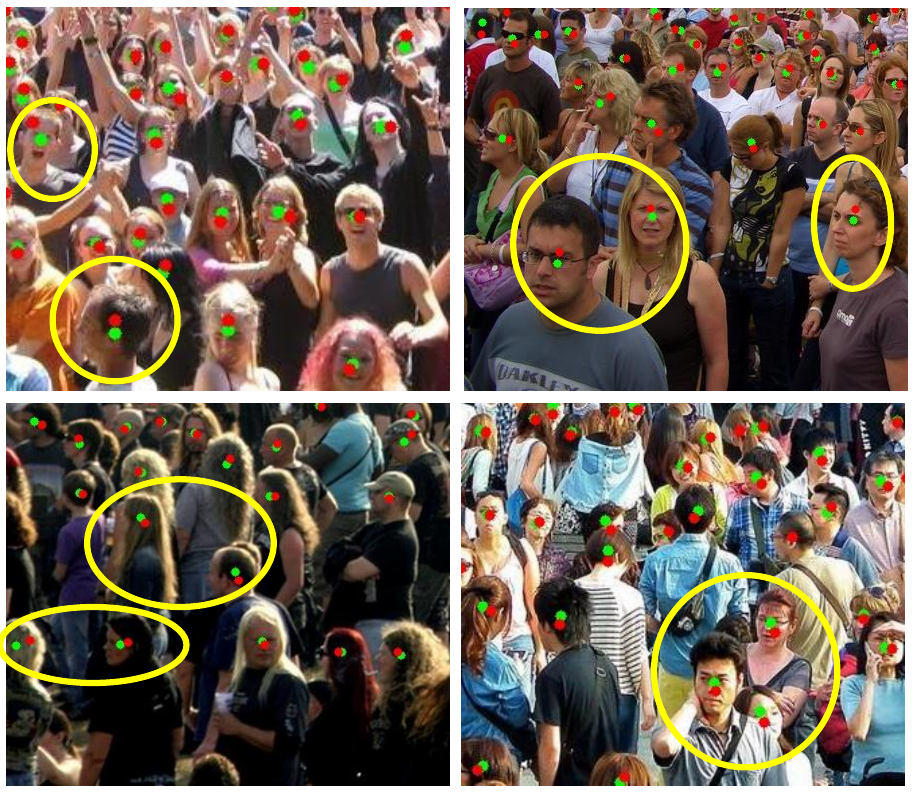}
	\caption{\textbf{Examples of point annotation refinement}. Red and green points are ground-truth points and refined points, respectively. The yellow circles highlight some refined points that significantly differ from the original ground-truth points. PET can distinguish the `noisy' ground-truth point and move them towards the center of heads.}
	\label{fig:anno_refine}
        \vspace{-10pt}
\end{figure}

\vspace{-5pt}
\paragraph{Application on Point Annotation Refinement.} 
Learning with noisy annotations~\cite{liu2022oamil,wan2020noisy} is a recent emerging topic.
Existing crowd counting benchmarks typically use a dot to represent a person. However, the noise may arise during the annotation process~\cite{wan2020noisy}, leading to inaccurate head points. In this context, another interesting application of PET is point annotation refinement. 
In practice, we first train PET with original annotations and validate the quality of annotations by setting annotated points as queries. 
Fig.~\ref{fig:anno_refine} shows some examples of the refined annotations. We observe that PET can distinguish `noisy' annotations and calibrate these points to the center of the head. To further investigate whether the refined annotations can benefit counting performance, we retrain PET with them. Our results show that the MAE on SH PartA improves from $49.34$ to $48.52$. 
While the improvement seems marginal at first glance, we remark that this is because most annotations are reasonable in the dataset. However, the visualizations in Fig.~\ref{fig:anno_refine} do imply the ability of PET for annotation refinement, which could make a difference in more challenging scenarios when accurate annotations are difficult to acquire.

\subsection{Ablation Study} \label{sec:ablation}

Here we justify the design choices of PET by conducting ablation studies on the ShanghaiTech PartA and UCF-QNRF datasets.

\begin{figure}[t]
    \centering
    \includegraphics[width=0.7\linewidth]{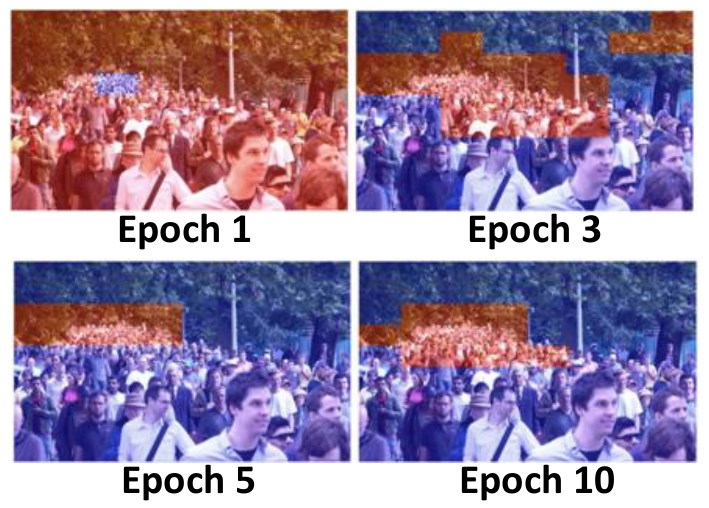}
    \captionof{figure}{Split maps of epoch $1,3,5,10$.}
    \label{fig:split_maps}
\end{figure}

\vspace{-5pt}
\paragraph{Learning Process of the Quadtree Splitter.} 
Fig.~\ref{fig:split_maps} shows the outputs of the quadtree splitter during training.
One can observe that the initial split map (epoch $1$) is meaningless. After a few training epochs, \eg, $10$ epochs, the quadtree splitter can output a reasonable split map. 
This suggests that Eq.~\eqref{eq:split} works well in supervising the quadtree splitter.

% Figure: split maps

% Ablation: point-query quadtree
\begin{table}[t]
\centering
\captionof{table}{Effect of point-query quadtree.}
\label{table:quadtree}
\setlength{\tabcolsep}{12.5pt}
\begin{tabular}[t]{@{}l c c@{}}
    \toprule
        Configuration & SH PartA & UCF-QNRF \\
        \midrule
        sparse point queries only & 53.59 & 90.70 \\
        dense point queries only & 54.16 & 98.14 \\
        point-query quadtree & \textbf{49.34} & \textbf{79.53} \\
        \bottomrule
    \end{tabular}
    \vspace{-10pt}
\end{table}

\vspace{-5pt}
\paragraph{Effect of Point-Query Quadtree.} 
Here we verify the effectiveness of the quadtree structure. 
For comparison, we train PET using only sparse point queries or dense point queries.
Table~\ref{table:quadtree} reports the results. 
Although sparse point queries can achieve promising results on SH PartA, it is far from satisfactory on UCF-QNRF.
In addition, the inferior results of dense point queries could be attributed to the ambiguity during bipartite matching, because a ground-truth point may correspond to several similar point queries when dealing with sparse regions. 
This impedes the model from discriminating valid points.

Comparatively, by adopting the point-query quadtree, we obtain 
notable improvements on both SH PartA ($\sim$4 MAE) and UCF-QNRF ($\sim$11 MAE). 
The quadtree structure ensures that sparse and dense regions can be processed by corresponding querying points, thus achieving better results.

% Ablation: Progressive Rectangle Window
\begin{table}[t]
    \centering
    \setlength{\tabcolsep}{5pt}
    \caption{Effect of progressive rectangle window attention. PET is trained with different configurations of attention. 
    }
    \begin{tabular}{@{} c l l l @{}}
      \toprule
      Progressive & Encoder Window & Decoder Window & MAE \\
      \midrule
      \checkmark & Rectangle & Square & 52.95\\
      \checkmark & Square & Rectangle & 51.93 \\
      \checkmark & Square & Square & 52.48\\
      \midrule
      \cross & Rectangle & Rectangle & 52.00 \\
      \checkmark & Rectangle & Rectangle & \textbf{49.34} \\
      \bottomrule
    \end{tabular}
    \label{table:rect_win}
\end{table}

% Ablation: Attention Comparison
\begin{table}[t]
    \centering
    \caption{Comparison of different attention mechanisms.}
    \begin{tabular}{@{}l | l  @{}}
        \toprule
        Attention Mechanism & MAE  \\
        \midrule
        Global Attention~\cite{nicolas2020detr} & 52.58  \\
        Deformable Attention~\cite{zhu2020deformable} & 51.0 \\
        \midrule
        Progressive Rectangle Window Attention (Ours) & \textbf{49.34} \\
        \bottomrule
    \end{tabular}
    \label{table:attention}
     \vspace{-10pt}
\end{table}

\vspace{-5pt}
\paragraph{Effect of Progressive Rectangle Window Attention.} 
To justify the effectiveness of the progressive rectangle window attention, we train with different configurations of PET as in Table~\ref{table:rect_win}. One can observe that:
i) progressive attention works, which significantly outperforms its non-progressive counterpart; 
ii) applying rectangle window to the encoder and decoder achieves the best performance.
We notice that replacing either encoder or decoder with square window deteriorates performance. 
The reason perhaps is that a rectangle window could capture more useful information about the crowd due to the perspective prior.

We also compare our attention with existing attention mechanisms. 
As shown in Table~\ref{table:attention}, while global attention and deformable attention obtain promising results, their performance fall behind our window attention. The inferior results of global attention could be attributed to the varied resolution of testing images, as the global-wise attention may not adapt well to arbitrary resolution. 
In addition, performing global attention can easily exceed the memory limit of the GPU when dealing with high-resolution images.

% Table: Efficiency
\begin{table}[t]
    \centering
    \setlength{\tabcolsep}{7pt}
    \caption{Comparison of parameters and inference time
    with localization-based methods. The inference time is measured on a NVIDIA 3090 GPU with $1024 \times 1024$ input.}
    \begin{tabular}{@{} l c c c @{}}
      \toprule
      Method & P2PNet~\cite{song2021p2p} & CLTR~\cite{liang2022cltr} & Ours \\
      \midrule
      Parameters (M) & 21.6 & 43.4 & \textbf{20.9} \\ 
      Inference Time (s) & \textbf{0.074} & 0.107 & 0.097\\
      \bottomrule
    \end{tabular}
    \label{table:efficiency}
\end{table}

% Figure: Encoder/Decoder Attention
\begin{figure}
     \centering
     \begin{subfigure}[b]{0.37\textwidth}
         \centering
         \includegraphics[width=\textwidth]{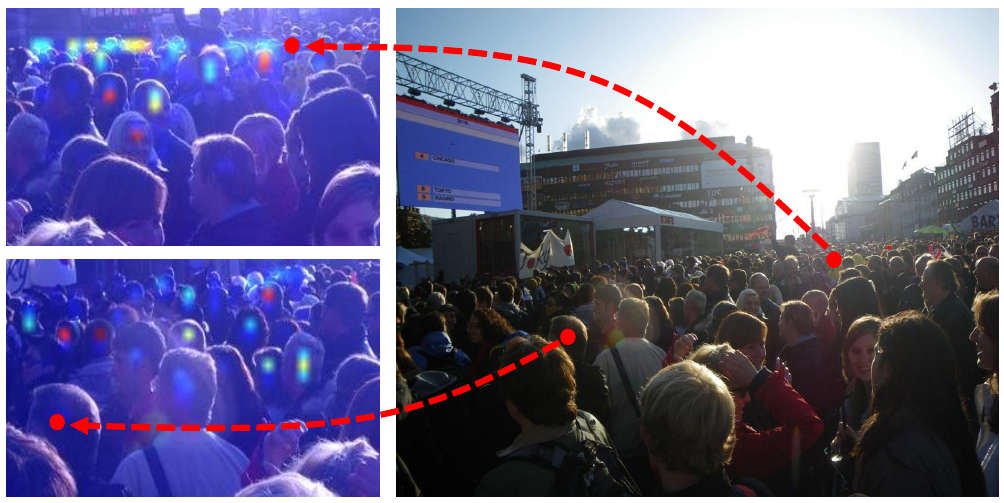}
         \caption{Encoder attention maps}
         \label{fig:enc_attention}
     \end{subfigure}
     \hfill
     \begin{subfigure}[b]{0.38\textwidth}
         \centering
         \includegraphics[width=\textwidth]{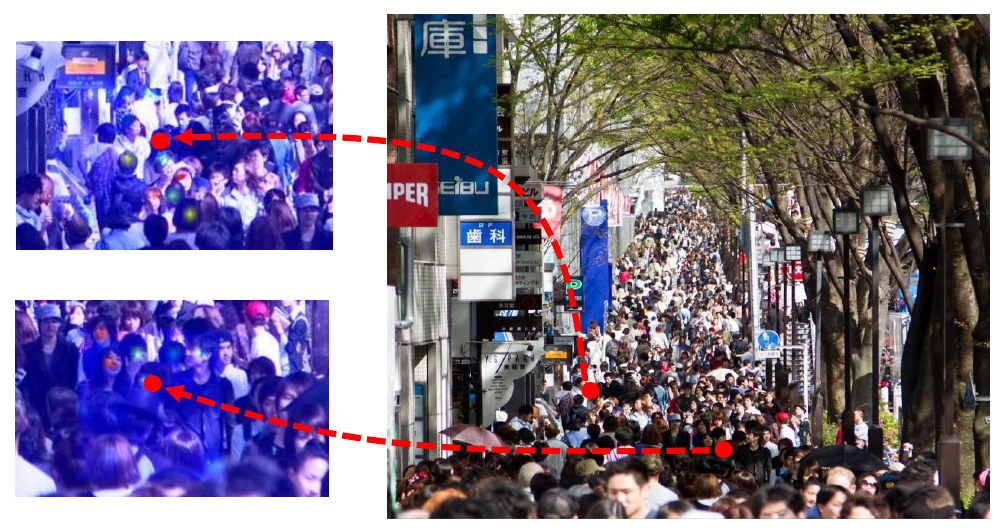}
         \caption{Decoder attention maps}
         \label{fig:dec_attention}
     \end{subfigure}
        \caption{\textbf{Visualization of encoder and decoder attention maps}. Red points in the input images are reference points.
	}
    \vspace{-10pt}
\end{figure}

\vspace{-5pt}
\paragraph{Efficiency of PET.} 
Table~\ref{table:efficiency} compares the model parameters and inference time of different models. Our PET owns the fewest parameters, and operates at a similar speed to existing methods. Note that the computational bottleneck lies in the backbone, instead of the querying process.

\vspace{-5pt}
\paragraph{Qualitative Results.} 
Here we show what is learned by the transformer encoder and decoder.
Fig.~\ref{fig:enc_attention} illustrates the self-attention maps of the transformer encoder for some reference points. The encoder can capture similar crowd within a rectangle window, thus can encode valuable context information. 
Fig.~\ref{fig:dec_attention} shows the decoder attention maps of two point queries. One can observe that higher attention value also occurs in similar crowd. 
Interestingly, the attention map of the encoder and decoder seems similar. 
This may attribute to the single-class nature of the crowd, \ie, the model only needs to distinguish human heads.

%%%%%%%%% CONCLUSION

\section{Conclusion}

We have shown that crowd counting can be viewed as a decomposable point querying process,
which provides an intuitive and universal way of modeling crowd. Specifically, we present a Point Query Transformer, featured by a point-query quadtree structure and the progressive rectangle window attention mechanism. Extensive experiments justify that PET implements the idea of decomposable point querying and exhibits generality in several crowd-related tasks. 

Albeit effective, PET still has limitations. For instance, it may suffer from missing detections when encountering large heads, because of the limited size of the rectangle window. The representation of the point query can also be improved.
For future work, we plan to extend our formulation to other dense prediction tasks. 

\vspace{10pt}
\noindent \textbf{Acknowledgement} This work was supported by the Natural Science Foundation of China under Grant No. U1913602, 61876211, and 62106080, and in part by the Chinese Fundamental Research Funds for the Central University under Grant 2021XXJS095.

{\small
\bibliographystyle{ieee_fullname}
\bibliography{egbib}
}

\end{document}